\documentclass{article}

\usepackage{PRIMEarxiv}

\usepackage[utf8]{inputenc} 
\usepackage[T1]{fontenc}    
\usepackage{hyperref}       
\usepackage{url}            
\usepackage{amsmath}        
\usepackage{amsfonts}       
\usepackage{microtype}      
\usepackage{fancyhdr}       
\usepackage{graphicx}       
\graphicspath{{figures/}{media/}} 

\newcommand{\singlefigurewidth}{0.74\linewidth}
\newcommand{\widefigurewidth}{0.84\textwidth}

\title{Wave Function Backpropagation with Explicit Temporal-Interval Dynamics}

\author{
  Byunggu Yu, Justin Kim \\
  Dept. of Computer Science and Information Technology \\
  University of the District of Columbia \\
  Washington, DC USA\\
  \texttt{\{byu, junwhan.kim\}@udc.edu} \\
}

\begin{document}
\maketitle

\begin{abstract}
Conventional neural networks learn predominantly through affine transformations followed by nonlinear activations, while elapsed time is often treated as an auxiliary feature or assumed to be uniformly sampled. This paper introduces Wave Function Backpropagation (WFB), a wave-parameterized learning formulation in which neural responses are represented by learnable amplitude, wavenumber, angular frequency, and phase. The formulation associates an observed state with its temporal interval $\Delta t$ through the phase of a differentiable spatiotemporal wave. We derive standard WFB gradients and a spatial-curvature correction based on the Laplacian of the wave response. WFB is instantiated in a deliberately feed-forward trajectory predictor to provide a controlled proof of concept; sequence learning is outside the scope of the present evaluation. With motion features, STD-WFB using real intervals reduces average displacement error (ADE) by 20.4\% relative to the original FFN baseline. In a new position-only evaluation that removes temporal leakage through precomputed velocity and acceleration, real-interval WFB reduces ADE by 10.4\% relative to the original FFN and remains competitive with parameter-matched ReLU controls, obtaining 2.1\% lower mean ADE than the matched FFN with explicit $\Delta t$. Shuffled-interval WFB attains the lowest mean ADE, indicating that the present evidence supports the effectiveness of the wave representation but does not attribute the gain to interval alignment. These results establish WFB as a viable structured feed-forward learning formulation and define a clear basis for subsequent architectural studies.
\end{abstract}


\section{Introduction}
Modern neural networks are largely built from affine transformations of the form $Wx+b$ followed by nonlinear activation functions~\cite{Rumelhart1986}. By composing these operations, deep networks can approximate complex input--output mappings and have achieved strong performance across vision, language, and sequential prediction. Nevertheless, the standard formulation does not provide a dedicated mechanism for representing elapsed time. In many applications, time is represented only by an observation index, appended as an ordinary feature, or absorbed indirectly into changes between consecutive states. These choices can be adequate under uniform sampling, but they provide no explicit parameterization of how a neural response should evolve when observation intervals vary.

Irregular temporal intervals arise in many settings, including asynchronous sensing, missed observations, event-based measurements, healthcare records, and physical motion. Existing approaches address this problem through time-dependent decay, continuous-time latent dynamics, or controlled differential equations~\cite{che2018grud,rubanova2019latentode,kidger2020neuralcde}. These methods have substantially advanced irregular time-series modeling. However, they do not define a general wave-parameterized learning rule in which an elapsed interval directly changes the phase of a learnable neural response.

This paper introduces Wave Function Backpropagation (WFB), a learning formulation that represents a neural component using amplitude, wavenumber, angular frequency, and phase. For spatial input $x$ and temporal interval $\Delta t$, the response is modeled through a phase term $kx-\omega\Delta t-\theta$. The interval therefore affects the activation through a structured phase displacement rather than only through feature concatenation. WFB remains trainable by gradient descent because all wave parameters are differentiable with respect to a task loss.

The purpose of this work is to establish the mathematical formulation and an initial empirical proof of concept for WFB. Trajectory prediction is used as the validation task rather than as the scope of the proposed framework. It is particularly suitable for this initial study because position changes and elapsed time have a direct physical relationship, and irregular frame gaps provide observable $\Delta t$ values. To isolate the effect of the proposed wave representation, we intentionally use a feed-forward network (FFN) rather than introducing recurrent memory, attention, or a task-specific interaction module. This controlled setting helps distinguish gains due to WFB from gains due to a stronger sequence architecture.

The contributions of this work are as follows:
\begin{itemize}
    \item We formulate WFB as a differentiable wave-parameterized learning mechanism in which spatial input and explicit temporal intervals interact through learnable amplitude, wavenumber, angular frequency, and phase.
    \item We derive standard task-gradient updates and a Laplacian-based curvature regularizer for the wave parameters, yielding standard, Laplacian-only, and combined standard--Laplacian variants.
    \item We provide a controlled feed-forward validation with real, shuffled, and constant $\Delta t$, position-only temporal controls, and parameter-matched ReLU baselines. These experiments separate the effectiveness of the complete WFB representation from model capacity and from temporal information embedded in engineered motion features.
\end{itemize}

The remainder of the paper is organized as follows. Section~\ref{rw} reviews related work. Section~\ref{math} develops the wave representation. Section~\ref{wfb} derives WFB and its curvature-corrected variants. Section~\ref{eval} presents the feed-forward proof-of-concept evaluation and discussion. Section~\ref{con} concludes the paper and identifies directions for future work.

\section{Related Work}\label{rw}

\subsection{Temporal Learning and Irregular Sampling}
Recurrent neural networks and Backpropagation Through Time model ordered observations through recursive hidden-state updates, but the recurrence index does not itself represent elapsed physical time. Time-aware extensions therefore modify hidden-state decay or gating using observation gaps. GRU-D, for example, uses trainable decay mechanisms to handle missing observations and irregular intervals~\cite{che2018grud}. Continuous-time approaches instead model latent dynamics between observations. Latent ODEs use ordinary differential equations to evolve hidden states over continuous time~\cite{rubanova2019latentode}, while Neural Controlled Differential Equations represent irregular sequences as continuous paths that drive latent dynamics~\cite{kidger2020neuralcde}. These methods focus on state evolution or interpolation. WFB is complementary: it investigates whether elapsed time can directly parameterize the phase of a learnable wave response.

Backpropagation Through Time, adjoint-based optimization, and biologically inspired continuous-time learning methods such as Generalized Latent Equilibrium address temporal credit assignment in recurrent or dynamical systems~\cite{Ellenberger2025}. Physics-Informed Neural Networks incorporate temporal variables through governing equations and residual constraints~\cite{raissi2019physics}. Spiking Neural Networks encode information through temporally structured spike events and are trained using methods such as surrogate gradients and spatio-temporal backpropagation~\cite{wu2018spatio,ma2025spikingneuralnetworkstemporal}. In contrast to these approaches, WFB defines the activation itself as a differentiable wave and optimizes its field parameters directly from a task loss.

\subsection{Periodic and Wave-Based Representations}
Fourier features and sinusoidal implicit representations have demonstrated that periodic bases can represent high-frequency or continuously varying signals effectively~\cite{tancik2020fourier,sitzmann2020implicit}. The present work differs in emphasis from fixed positional encodings or a generic sine activation: amplitude, spatial frequency, temporal frequency, and phase are exposed as learnable parameters of the same response, and an observed interval directly enters its phase. A Laplacian-derived term is additionally studied as a selected-parameter curvature correction on the learned wave field.

\subsection{Trajectory Prediction as a Validation Task}
Trajectory prediction estimates future coordinates from an observed history of agent positions. Social-LSTM introduced social pooling for pedestrian interactions, and Social-GAN modeled multimodal socially acceptable futures~\cite{7780479,gupta2018socialgan}. More recent trajectory models emphasize interaction modeling, multimodality, attention, and motion priors~\cite{wang2024mdstf,zhang2024pedestrian,song2024motion,enhanced2024state,bae2024singulartrajectory}. These task-specific architectures are not the focus of the present study. Instead, trajectory prediction provides a physically interpretable testbed for evaluating whether a wave-parameterized activation can learn from spatial states and observed temporal intervals under a deliberately controlled FFN architecture.

\section{Wave-Parameterized Representation}\label{math}

\subsection{From Static Activations to Spatiotemporal Waves}
A conventional single-hidden-layer approximation can be written as
\begin{equation}
F(x)=\sum_{n=1}^{N}a_n\,\sigma(w_nx+b_n),
\end{equation}
where each component applies an affine transformation followed by an activation function. WFB instead represents each component as a parameterized harmonic response. For a scalar spatial input $x$ and elapsed temporal interval $\Delta t$, the complex wave is
\begin{equation}
\begin{aligned}
\psi_n(x,\Delta t) &= A_n e^{i\phi_n},\\
\phi_n &= k_nx-\omega_n\Delta t-\theta_n,
\end{aligned}
\label{eq:complex_wave}
\end{equation}
where $A_n$ is amplitude, $k_n$ is wavenumber, $\omega_n$ is angular frequency, and $\theta_n$ is phase. The real-valued response used by the network is
\begin{equation}
f_n(x,\Delta t)=A_n\cos(\phi_n).
\label{eq:real_wave}
\end{equation}
A wave layer with $N$ components produces
\begin{equation}
F(x,\Delta t)=\sum_{n=1}^{N}f_n(x,\Delta t).
\end{equation}
For vector inputs, $k_nx$ is replaced by an inner product $\mathbf{k}_n^{\top}\mathbf{x}$. Thus,
\begin{equation}
f_n(\mathbf{x},\Delta t)
  =A_n\cos\!\left(\mathbf{k}_n^{\top}\mathbf{x}
  -\omega_n\Delta t-\theta_n\right).
\label{eq:vector_wave}
\end{equation}
Equation~\eqref{eq:vector_wave} is the practical formulation used in a neural layer. It directly associates the state $\mathbf{x}$ with the elapsed interval $\Delta t$ through phase. Importantly, the temporal interval is not claimed to encode sequence order by itself; it represents elapsed time associated with the current observation. Learning ordered temporal dependencies requires an additional propagation mechanism, which is left for future recurrent or attention-based WFB architectures.

\subsection{Separable Spatial and Temporal Amplitudes}
The implementation evaluated in this paper uses a separable amplitude
\begin{equation}
A_n=A_{xn}A_{tn},
\end{equation}
which allows spatial and temporal contributions to be optimized separately. The component response becomes
\begin{equation}
\begin{aligned}
f_n(x,\Delta t) &= A_{xn}A_{tn}\cos(\phi_n),\\
\phi_n &= k_nx-\omega_n\Delta t-\theta_n.
\end{aligned}
\label{eq:separable_wave}
\end{equation}
This factorization is an architectural choice rather than a requirement of WFB. A single amplitude or vector-valued amplitudes may be used in other implementations.

\subsection{Learnable Parameters}
Unlike a conventional layer that primarily optimizes weights and biases, a WFB component optimizes the field parameters
\begin{equation}
\Theta_n=\{A_{xn},A_{tn},k_n,\omega_n,\theta_n\}.
\end{equation}
The amplitude determines response magnitude, $k_n$ determines sensitivity to the spatial input, $\omega_n$ determines phase change per unit elapsed time, and $\theta_n$ determines phase offset. These parameters are learned jointly with the remaining network parameters using the task objective.

\section{Wave Function Backpropagation}\label{wfb}
Let $\mathcal{L}$ denote the task loss and define the upstream derivative
\begin{equation}
\delta_n=\frac{\partial\mathcal{L}}{\partial f_n}.
\end{equation}
For clarity, the following derivation uses the scalar form in Eq.~\eqref{eq:separable_wave}; the vector form follows by replacing $x$ with $\mathbf{x}$ and $k_n$ with $\mathbf{k}_n$.

\subsection{Standard WFB Gradients}
The partial derivatives of the wave response are
\begin{align}
\frac{\partial f_n}{\partial A_{xn}}
&=A_{tn}\cos(\phi_n),\\
\frac{\partial f_n}{\partial A_{tn}}
&=A_{xn}\cos(\phi_n),\\
\frac{\partial f_n}{\partial k_n}
&=-A_{xn}A_{tn}x\sin(\phi_n),\\
\frac{\partial f_n}{\partial \omega_n}
&=A_{xn}A_{tn}\Delta t\sin(\phi_n),\\
\frac{\partial f_n}{\partial \theta_n}
&=A_{xn}A_{tn}\sin(\phi_n).
\end{align}
Applying the chain rule gives
\begin{align}
\frac{\partial\mathcal{L}}{\partial A_{xn}}
&=\delta_nA_{tn}\cos(\phi_n),\\
\frac{\partial\mathcal{L}}{\partial A_{tn}}
&=\delta_nA_{xn}\cos(\phi_n),\\
\frac{\partial\mathcal{L}}{\partial k_n}
&=-\delta_nA_{xn}A_{tn}x\sin(\phi_n),\\
\frac{\partial\mathcal{L}}{\partial \omega_n}
&=\delta_nA_{xn}A_{tn}\Delta t\sin(\phi_n),\\
\frac{\partial\mathcal{L}}{\partial \theta_n}
&=\delta_nA_{xn}A_{tn}\sin(\phi_n).
\end{align}
These equations define standard WFB. The $\omega_n$ gradient is explicitly scaled by $\Delta t$, causing observations separated by different elapsed intervals to produce different temporal-frequency updates.

\subsection{Laplacian Curvature Regularization}
To study whether curvature control stabilizes the learned wave field, we introduce a penalty based on the spatial Laplacian. In one spatial dimension,
\begin{align}
\frac{\partial f_n}{\partial x}
&=-k_nA_{xn}A_{tn}\sin(\phi_n),\\
\Delta_x f_n
=\frac{\partial^2f_n}{\partial x^2}
&=-k_n^2A_{xn}A_{tn}\cos(\phi_n)
=-k_n^2f_n.
\end{align}
The curvature penalty is
\begin{equation}
\begin{aligned}
\mathcal{P}
  &=\frac{\lambda}{2}\sum_{n=1}^{N}(\Delta_x f_n)^2\\
  &=\frac{\lambda}{2}\sum_{n=1}^{N}k_n^4f_n^2,
\end{aligned}
\label{eq:penalty}
\end{equation}
where $\lambda\geq0$ controls regularization strength. For parameters $q\in\{A_{tn},\omega_n,\theta_n\}$, while treating $k_n$ as fixed in this correction branch,
\begin{equation}
\frac{\partial\mathcal{P}}{\partial q}
=\lambda k_n^4f_n\frac{\partial f_n}{\partial q}.
\label{eq:penalty_grad}
\end{equation}
Substitution into Eq.~\eqref{eq:penalty_grad} gives the three correction components:
\begin{align}
\frac{\partial\mathcal{P}}{\partial A_{tn}}
  &=\lambda k_n^4 A_{xn}^2A_{tn}\cos^2(\phi_n),\\
\frac{\partial\mathcal{P}}{\partial \omega_n}
  &=\lambda k_n^4\Delta t\,A_{xn}^2A_{tn}^2
    \cos(\phi_n)\sin(\phi_n),\\
\frac{\partial\mathcal{P}}{\partial \theta_n}
  &=\lambda k_n^4A_{xn}^2A_{tn}^2
    \cos(\phi_n)\sin(\phi_n).
\label{eq:penalty_components}
\end{align}

Because Eq.~\eqref{eq:penalty} is derived from spatial curvature, it should not be interpreted as a temporal Laplacian. Its effect on $\omega_n$ and $\theta_n$ arises because these parameters also control the same spatiotemporal phase. The $k_n^4$ factor increasingly penalizes high-spatial-frequency responses.

\subsection{Combined Standard--Laplacian WFB}
The implementation treats the Laplacian term as a custom correction for the selected temporal parameters
$\mathcal{Q}_t=\{A_{tn},\omega_n,\theta_n\}$ rather than as a global loss differentiated through every network parameter. For $q\in\mathcal{Q}_t$, the two correction modes are
\begin{align}
g_q^{\mathrm{lap}}
  &=\frac{\partial\mathcal{P}}{\partial q},\\
g_q^{\mathrm{comb}}
  &=\frac{\partial\mathcal{L}}{\partial q}
    +\frac{\partial\mathcal{P}}{\partial q}.
\end{align}
The wavenumber is held fixed in this correction branch, and all parameters outside $\mathcal{Q}_t$ continue to receive their supervised task gradients. Standard WFB uses only $\partial\mathcal{L}/\partial q$; Laplacian-only WFB replaces the supervised gradients of the selected temporal parameters with $g_q^{\mathrm{lap}}$; and combined standard--Laplacian WFB uses $g_q^{\mathrm{comb}}$. This formulation matches the implemented update and avoids interpreting the selected-parameter correction as optimization of a global objective $\mathcal{L}+\mathcal{P}$.

\section{Proof-of-Concept Evaluation}\label{eval}

\subsection{Evaluation Scope}
This proof-of-concept evaluation addresses three questions. First, does inserting a wave-parameterized representation into a feed-forward trajectory predictor improve accuracy relative to conventional FFNs? Second, does WFB remain competitive when temporal leakage through engineered velocity and acceleration is removed and model capacity is controlled? Third, how do the standard, Laplacian-only, and combined standard--Laplacian update rules behave as the correction weight $\lambda$ changes? All experiments use feed-forward predictors to isolate the WFB representation. Sequence learning, recurrent state propagation, attention, and comparison with specialized trajectory-prediction systems are intentionally outside the scope of this study.

\subsection{Data Preparation and Temporal Intervals}
We use ETH/UCY and JAAD-style pedestrian and agent annotations~\cite{kotseruba2016jaad,lerner2007crowd}. Frame-level detections are associated across frames to form pseudo-trajectories. To prevent overlapping windows from the same agent from appearing in different partitions, the data are split by the pair $({\rm source},{\rm agent\_id})$ before normalization and window extraction. The resulting irregularly sampled set contains 425,090 frame-level observations from 14,703 trajectories and 453,338 windows: 322,160 for training, 61,794 for validation, and 69,384 for testing.

For observation $i$, the elapsed time is computed before normalization as
\begin{equation}
    \Delta t_i = \frac{F_i-F_{i-1}}{\mathrm{FPS}},
    \label{eq:eval_dt}
\end{equation}
where $F_i$ and $F_{i-1}$ are consecutive available frame indices for the same trajectory and $\mathrm{FPS}=2.5$. Thus, missed detections and nonuniform frame gaps remain visible to the model. The extracted intervals have mean $0.842$, standard deviation $0.551$, and range $[0.398,10.801]$ seconds. Both the motion features and $\Delta t$ are standardized using statistics estimated from the training partition only.

Each observation is represented by
\begin{equation}
    \mathbf{x}_i = [x_i,y_i,v_i^x,v_i^y,a_i^x,a_i^y],
\end{equation}
where $(x_i,y_i)$ is the normalized bounding-box center and
\begin{align}
    v_i^x &= \frac{x_i-x_{i-1}}{\Delta t_i}, &
    v_i^y &= \frac{y_i-y_{i-1}}{\Delta t_i},\\
    a_i^x &= \frac{v_i^x-v_{i-1}^x}{\Delta t_i}, &
    a_i^y &= \frac{v_i^y-v_{i-1}^y}{\Delta t_i}.
\end{align}
An input window contains $T_{\rm obs}=8$ observations and the target contains the next $T_{\rm pred}=12$ center coordinates. Since the source annotations provide normalized image coordinates rather than world coordinates, ADE, FDE, MSE, and RMSE are reported in normalized coordinate units, not meters.

\subsection{Models and Training Protocol}
The original FFN baseline flattens the $8\times6$ motion-feature matrix and maps it directly to $12\times2$ future coordinates. It does not receive $\Delta t$ as a separate variable. WFB-FFN first projects each observation to a 128-dimensional hidden state and applies the wave response in Eq.~\eqref{eq:vector_wave}; the resulting eight wave states are flattened and passed to the same type of feed-forward prediction head. The WFB block therefore provides explicit amplitude, wavenumber, angular-frequency, and phase parameters, while $\Delta t_i$ modulates the phase at each observed position.

The new capacity-controlled experiment uses only position $(x,y)$ as the spatial input, preventing real-$\Delta t$ information from entering through precomputed velocity or acceleration. It compares the original FFN, an FFN with explicit $\Delta t$, parameter-matched versions of both FFNs, and WFB with real, shuffled, or constant intervals. The original position-only FFN has 142,104 trainable parameters; the matched FFN, matched FFN with $\Delta t$, and WFB have 400,753, 402,452, and 401,560 parameters, respectively. A parameter-matched sinusoidal FFN was also run as an optimization control, but its shared hyperparameter setting was unstable; it is therefore excluded from the efficacy comparison.

We evaluate three WFB update rules. STD-WFB-FFN uses the task-loss gradient for all parameters. Laplacian-WFB-FFN replaces the standard gradients of the temporal parameters $(A_t,\omega,\theta_t)$ with the curvature-derived gradients scaled by $\lambda$. STD-Laplacian-WFB-FFN instead adds the scaled curvature-derived gradients to their standard task gradients. Thus, $\lambda$ weights a temporal gradient correction; it is not an additional term in the reported MSE objective.

All models are optimized with MSE loss and AdamW using a learning rate of $10^{-3}$, weight decay of $10^{-4}$, batch size 512, gradient clipping at 1.0, and a maximum of 100 epochs. Early stopping uses validation ADE with patience 15. The principal comparisons use the same five seeds. Results are reported as mean $\pm$ standard deviation; with only five runs, small differences are interpreted descriptively rather than as definitive statistical superiority.

For a predicted trajectory $\hat{\mathbf{p}}_{1:T_{\rm pred}}$ and ground truth $\mathbf{p}_{1:T_{\rm pred}}$, the principal metrics are
\begin{align}
    \mathrm{ADE} &= \frac{1}{T_{\rm pred}}\sum_{j=1}^{T_{\rm pred}}
    \left\lVert \hat{\mathbf{p}}_j-\mathbf{p}_j \right\rVert_2,\\
    \mathrm{FDE} &= \left\lVert \hat{\mathbf{p}}_{T_{\rm pred}}-
    \mathbf{p}_{T_{\rm pred}} \right\rVert_2.
\end{align}
We additionally report coordinate-wise MSE and its square root (RMSE). Lower values are better for all metrics.

Unless otherwise stated, all experiments use the same ReLU-based MLP trajectory decoder to ensure a consistent comparison across WFB variants and the FFN baseline. The linear decoder introduced in Section~\ref{decoder} is evaluated only as an architectural ablation to investigate whether the nonlinear wave representation can reduce the need for an additional nonlinear decoding network.

\subsection{Trajectory Prediction and Temporal Ablation}
\begin{figure}[t]
    \centering
    \includegraphics[width=\widefigurewidth]{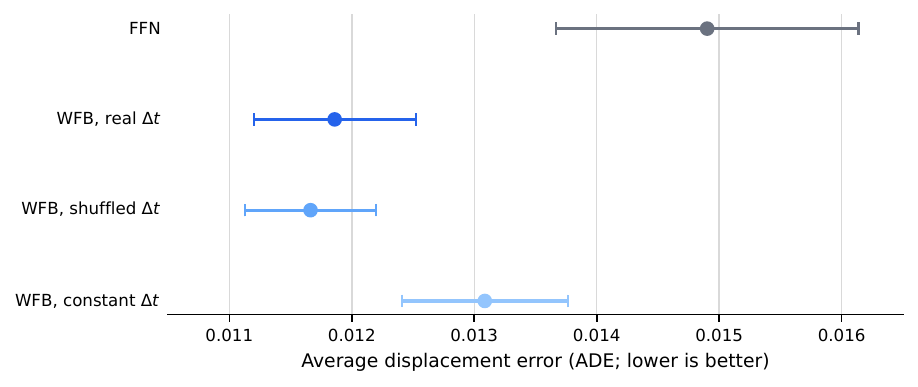}
    \caption{Motion-feature trajectory prediction under temporal-input ablations. Points show mean ADE and error bars show standard deviation over five seeds. Lower is better.}
    \label{fig:motion_temporal_ablation}
\end{figure}

Figure~\ref{fig:motion_temporal_ablation} shows that STD-WFB-FFN with real intervals reduces mean ADE by $20.4\%$ relative to the original FFN baseline; FDE, MSE, and RMSE improve in the same direction. The constant-$\Delta t$ model still reduces ADE by $12.2\%$, showing that a substantial part of the gain is associated with the complete WFB parameterization rather than temporal variation alone. Because the WFB block adds a projection and wave parameters, this comparison does not by itself isolate the wave representation from increased model capacity; the position-only experiment below addresses that question directly.

The shuffled-$\Delta t$ condition attains the lowest mean error, improving ADE by $21.7\%$ relative to FFN and slightly outperforming real $\Delta t$. Shuffling preserves the interval distribution while breaking its observation-level alignment. Accordingly, this feed-forward proof of concept supports the effectiveness of the WFB representation but does not attribute its gain to correct interval alignment, which is not a claim tested in the present study.

\subsection{Position-Only Capacity-Controlled Evaluation}
\begin{figure}[t]
    \centering
    \includegraphics[width=\widefigurewidth]{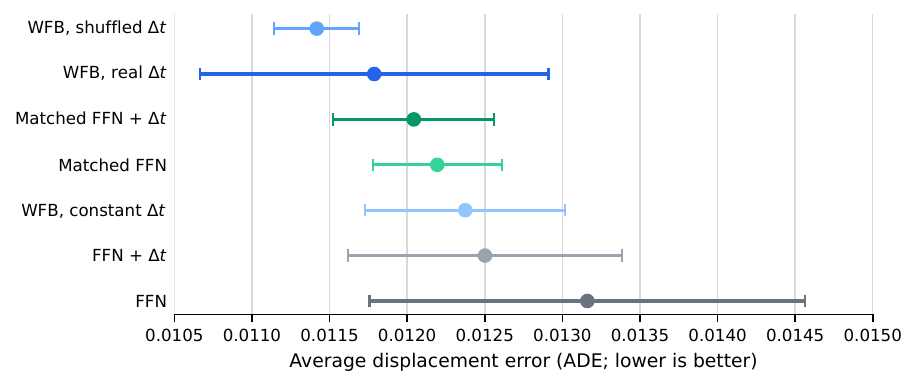}
    \caption{Position-only capacity-controlled comparison. Points show mean ADE and error bars show standard deviation over five seeds. The matched ReLU controls have approximately the same number of trainable parameters as WFB.}
    \label{fig:position_control_ade}
\end{figure}

\begin{figure}[t]
    \centering
    \includegraphics[width=\widefigurewidth]{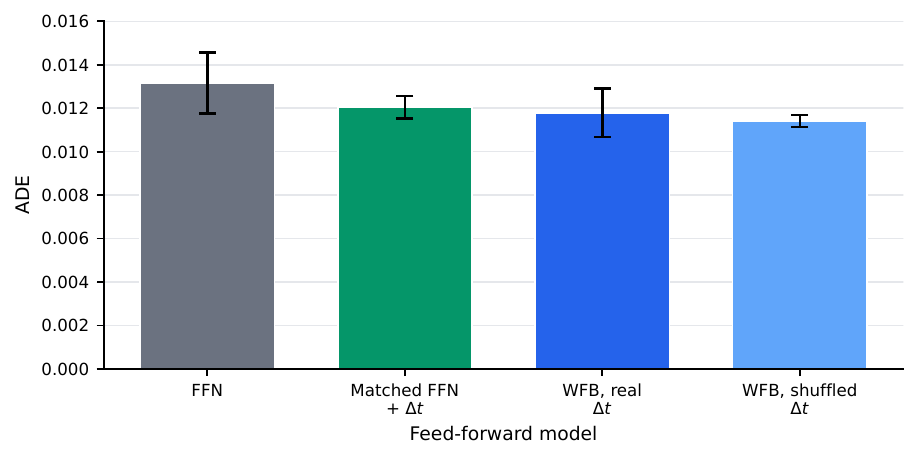}
    \caption{Position-only ADE for four principal feed-forward controls. Bars show the five-seed mean and error bars show one standard deviation. Lower is better.}
    \label{fig:position_control_seeds}
\end{figure}

Figure~\ref{fig:position_control_ade} reports the new evaluation using only observed positions as spatial inputs. Real-interval WFB obtains an ADE of $0.011787\pm0.001123$, improving on the original position-only FFN by $10.4\%$. More importantly, it remains competitive after controlling capacity: its mean ADE is $2.1\%$ lower than the parameter-matched FFN with explicit $\Delta t$ ($0.012043\pm0.000518$) and $3.3\%$ lower than the parameter-matched FFN without $\Delta t$ ($0.012194\pm0.000416$). These differences are modest relative to five-seed variability, so the result is evidence of competitive effectiveness rather than definitive superiority.

Shuffled-interval WFB achieves the lowest mean ADE, $0.011418\pm0.000274$, whereas constant-interval WFB obtains $0.012374\pm0.000644$. The four-model bar chart in Figure~\ref{fig:position_control_seeds} highlights the comparison among the original FFN, the strongest matched FFN control with explicit $\Delta t$, and WFB with real or shuffled intervals. Together, these results support two bounded conclusions: WFB is an effective feed-forward representation under both motion-feature and position-only inputs, and correct interval alignment is not established as the source of the improvement in this experiment.

\subsection{Backpropagation Rule and Laplacian Weight}
\begin{figure}[t]
    \centering
    \includegraphics[width=\singlefigurewidth]{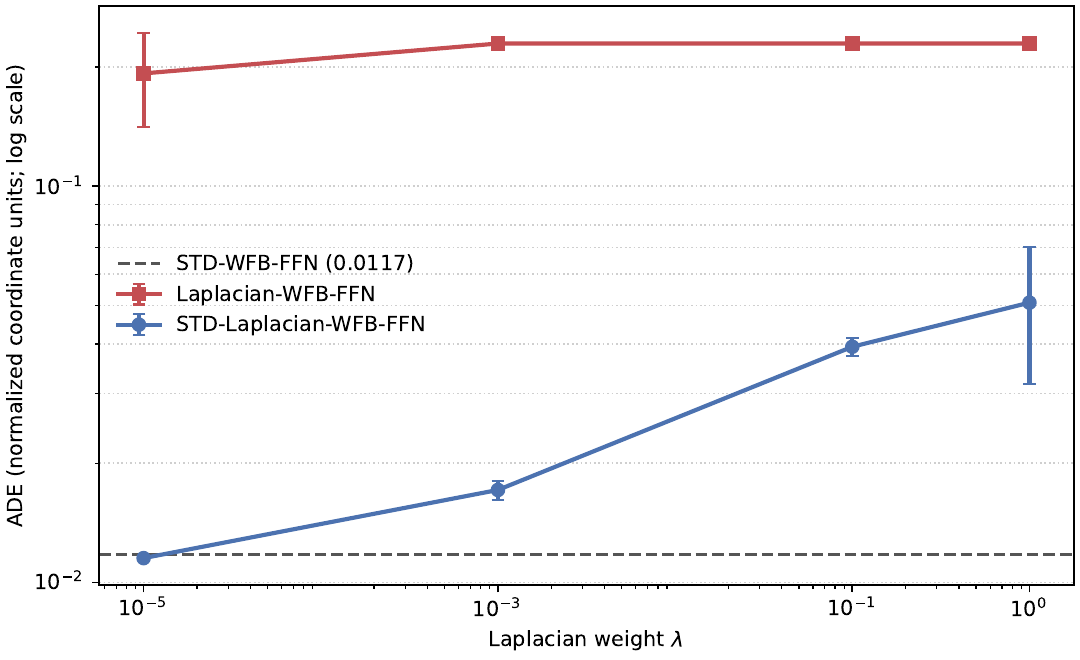}
    \caption{ADE as a function of the Laplacian gradient weight $\lambda$ under real $\Delta t$. Error bars show standard deviation across runs. The ADE axis is logarithmic because Laplacian-only WFB is more than an order of magnitude worse than the supervised variants.}
    \label{fig:lambda_sweep}
\end{figure}

Figure~\ref{fig:lambda_sweep} separates the effect of the update rule from the effect of $\lambda$. STD-WFB-FFN obtains an ADE of $0.011739\pm0.000640$. The best combined setting, STD-Laplacian-WFB-FFN with $\lambda=10^{-5}$, obtains $0.011513\pm0.000039$, a modest $1.9\%$ reduction. Increasing $\lambda$ to $10^{-3}$, $10^{-1}$, and $10^{0}$ increases ADE to $0.017112$, $0.039341$, and $0.050845$, respectively. The correction is therefore useful only when it remains weak relative to the supervised gradient; the results do not support the claim that a larger Laplacian contribution is better.

Laplacian-only WFB performs poorly for every tested weight, with ADE between $0.192761$ and $0.229382$. Its temporal update is determined by local wave curvature rather than by the direction that minimizes trajectory displacement. Applied alone, this update can suppress or redirect the temporal parameters without providing sufficient task-level credit assignment. The combined rule avoids this failure because it retains the supervised gradient and uses curvature only as a small correction. The resulting $1.9\%$ improvement over standard WFB indicates that curvature information is most effective as a complementary learning signal.

\subsection{Interpretability Analysis}
\begin{figure*}[t]
    \centering
    \includegraphics[width=\widefigurewidth]{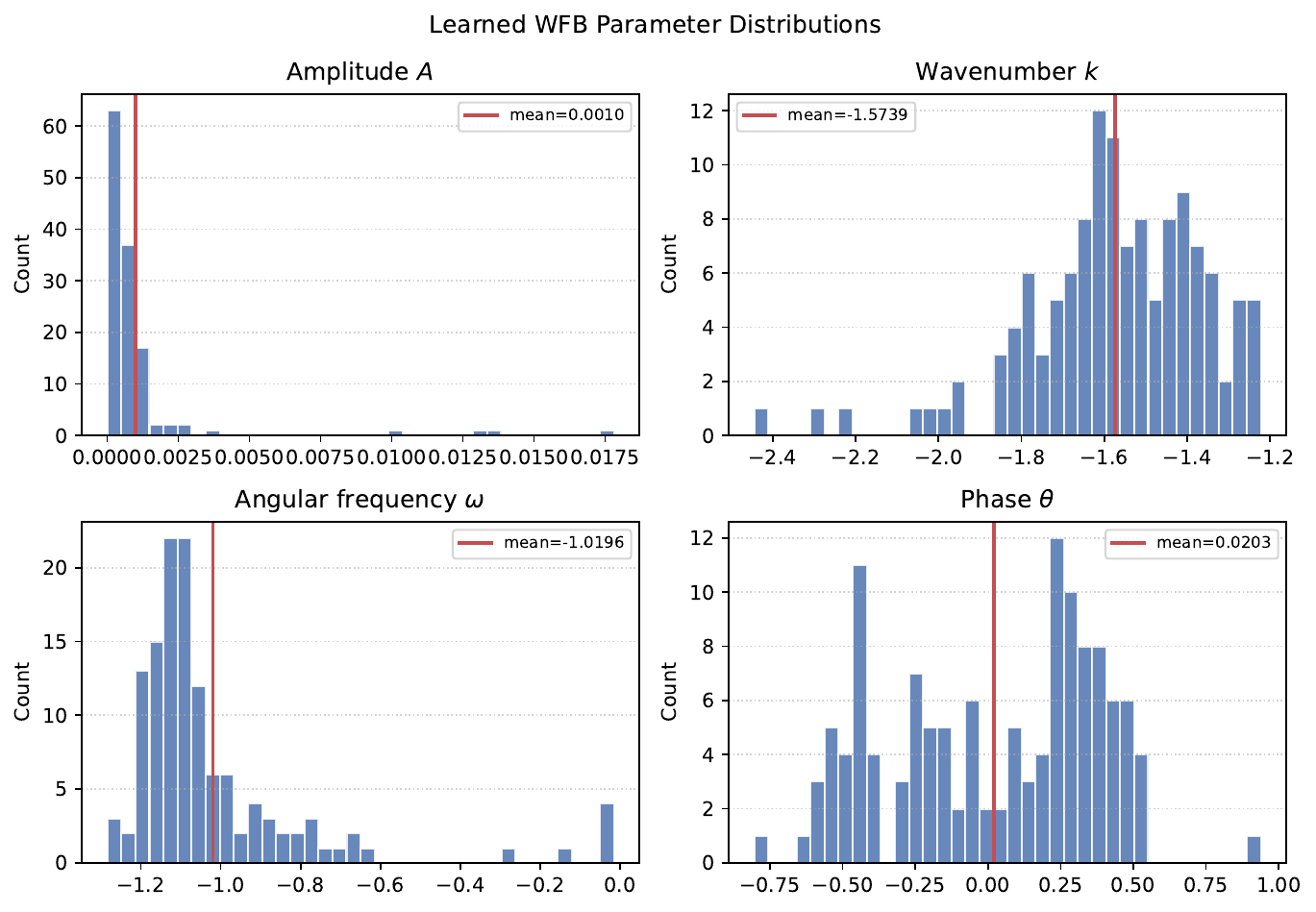}
    \caption{Distributions of the learned effective amplitude $A=A_xA_t$, wavenumber $k$, angular frequency $\omega$, and phase $\theta=\theta_x+\theta_t$ for the analyzed STD-Laplacian-WFB-FFN checkpoint.}
    \label{fig:wave_parameter_distributions}
\end{figure*}

Figure~\ref{fig:wave_parameter_distributions} shows that the trained model does not collapse all wave dimensions to a common value. Across the 128 wave channels, the effective amplitude is sparse and right-skewed ($0.0010\pm0.0024$), whereas $k$ and $\omega$ exhibit broader channel-dependent distributions ($-1.5739\pm0.2138$ and $-1.0196\pm0.2445$, respectively). The phase distribution spans both signs ($0.0203\pm0.3602$). These statistics establish parameter diversity, but they should not be interpreted as physical frequencies or wavelengths because the inputs and hidden states are normalized and the signs of the learned parameters are unconstrained.

To examine whether this diversity is related to trajectory structure, we compute Spearman correlations between activation-weighted wave parameters and sample-level motion descriptors. The strongest associations are with mean speed: $\rho=-0.322$ for $A$, $-0.283$ for $k$, $-0.297$ for $\omega$, and $0.300$ for $\theta$. Similar, moderate associations occur for path length, displacement, and mean acceleration. In contrast, correlations with the mean, standard deviation, and maximum of $\Delta t$ are mostly weak; the largest is $\rho=-0.197$ between $\theta$ and mean $\Delta t$. Thus, the learned wave representation is measurably associated with motion regime, while direct encoding of interval statistics is comparatively limited. Consistently, activation-weighted $|\omega|$ increases from approximately $0.639$ for slow motion to $0.678$ for fast motion, although the distributions overlap substantially.

\begin{figure}[t]
    \centering
    \includegraphics[width=\singlefigurewidth]{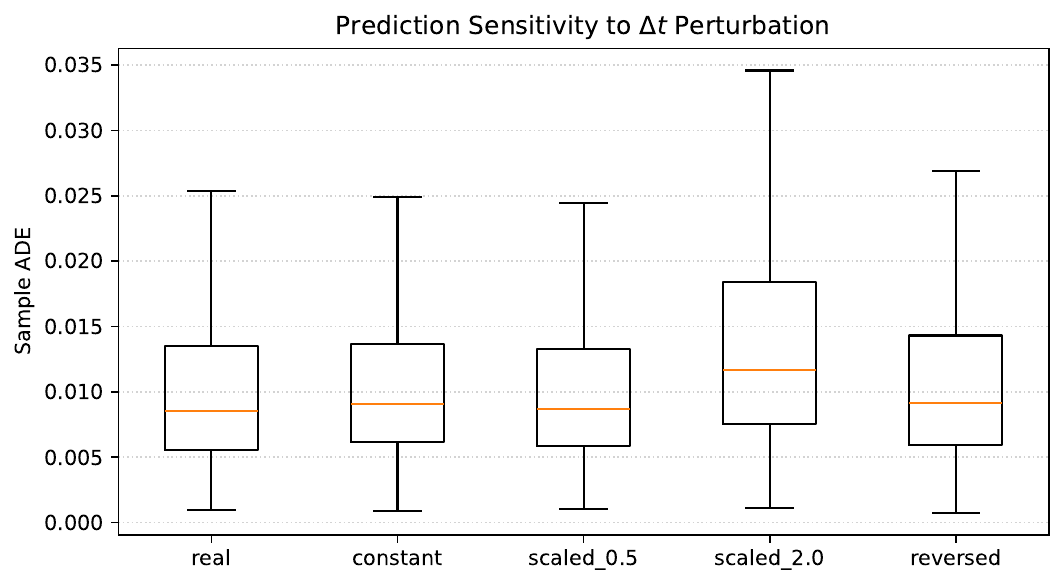}
    \caption{Test-sample ADE after intervening on $\Delta t$ while holding the trained model and spatial/motion inputs fixed. The analysis uses 5,120 samples from one STD-Laplacian-WFB-FFN checkpoint.}
    \label{fig:dt_sensitivity}
\end{figure}

The intervention in Figure~\ref{fig:dt_sensitivity} complements the training ablation by changing only $\Delta t$ at inference. Mean ADE is $0.011210$ with real intervals, $0.012017$ with constant intervals, $0.012198$ with reversed intervals, $0.011434$ with intervals scaled by $0.5$, and $0.014672$ with intervals scaled by $2.0$. Relative to real $\Delta t$, these changes correspond to increases of $7.2\%$, $8.8\%$, $2.0\%$, and $30.9\%$, respectively. The fixed model is therefore sensitive to temporal-interval values, especially to large changes in scale. However, sensitivity alone does not prove that the model has learned the correct interval ordering, which remains consistent with the real-versus-shuffled training result.

\subsection{Decoder Ablation and Parameter Efficiency}\label{decoder}

\begin{figure}[t]
    \centering
    \includegraphics[width=\widefigurewidth]{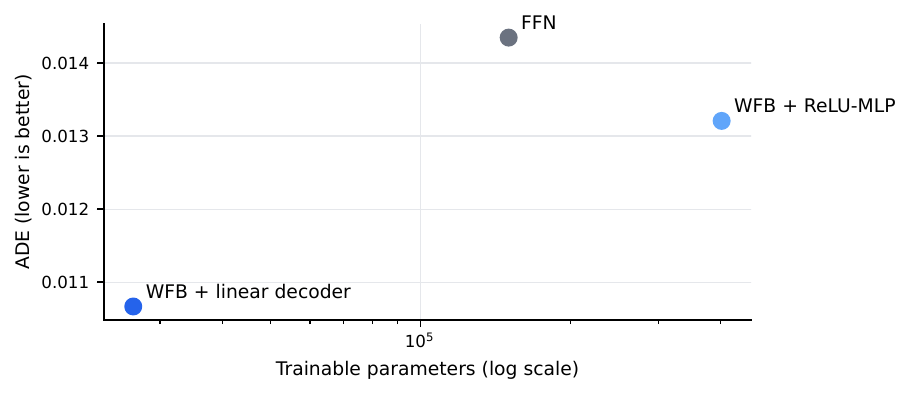}
    \caption{Accuracy--parameter trade-off in the decoder ablation. ADE is plotted against the number of trainable parameters on a logarithmic scale; lower ADE and fewer parameters are better.}
    \label{fig:decoder_efficiency}
\end{figure}

A conventional multilayer FFN requires nonlinear activations such as ReLU between affine layers; otherwise, multiple affine transformations collapse into a single affine mapping. In contrast, STD-WFB introduces nonlinearity directly through the phase-dependent wave response $A_xA_t\cos(kz-\omega\Delta t-\theta)$. We therefore evaluate whether the ReLU-based MLP trajectory decoder can be removed and replaced with a single linear output layer without degrading prediction accuracy.

As shown in Figure~\ref{fig:decoder_efficiency}, removing the ReLU-MLP decoder does not reduce the predictive accuracy of STD-WFB. Instead, the linear-decoder STD-WFB achieves the lowest ADE in this ablation, reducing ADE by $25.7\%$ relative to the FFN baseline and by $19.2\%$ relative to STD-WFB with the ReLU-MLP decoder. This result suggests that the nonlinear wave representation already captures much of the structure required for trajectory prediction, allowing a lightweight linear layer to decode the learned features without additional nonlinear transformations.

The decoder removal also substantially reduces model complexity. The parameter count decreases from 402,072 to 26,520, corresponding to a $93.4\%$ reduction relative to the original STD-WFB predictor and an $82.4\%$ reduction relative to the FFN baseline. The number of linear operations is reduced from 405,504 to approximately 30,720 MACs per sample, although the WFB layer additionally evaluates 1,024 cosine responses per sample.

Overall, these results suggest that the nonlinear expressiveness of STD-WFB is primarily provided by the wave representation itself rather than by the decoder. Consequently, the ReLU-based MLP decoder can be replaced with a lightweight linear readout without sacrificing predictive accuracy in this controlled setting. This should not be interpreted as evidence that linear decoders are universally superior; instead, it indicates that the proposed wave representation substantially reduces the need for an additional nonlinear decoding network.

\subsection{Discussion}
A wave representation is useful for trajectory prediction because it provides a continuous function of both spatial state and elapsed time. A conventional FFN approximates continuously evolving motion by composing affine transformations and pointwise activations, which partition the input space into piecewise-linear regions. WFB instead couples state and $\Delta t$ within the same phase function. Its amplitude controls response strength, $k$ controls spatial variation, $\omega$ controls temporal variation, and $\theta$ controls alignment. These variables are trained simultaneously within one wave response, allowing multiple aspects of trajectory dynamics to interact directly rather than being represented only through additional linear partitions.

The evaluation supports the effectiveness of this formulation from complementary perspectives. With motion features, STD-WFB-FFN using real $\Delta t$ reduces ADE by $20.4\%$ relative to the original FFN, while the constant-$\Delta t$ variant improves by $12.2\%$. With position-only inputs, real-interval WFB improves ADE by $10.4\%$ relative to the original FFN and remains competitive with parameter-matched ReLU controls. This capacity-controlled result is important: it shows that WFB's performance cannot be explained solely by comparison with a smaller baseline, while the modest differences among the matched models appropriately bound the strength of the claim.

Inference-time interventions further confirm that $\Delta t$ is operational rather than merely appended to the feature vector: replacing, reversing, or rescaling the intervals changes the predictions, with doubling the intervals increasing ADE by $30.9\%$. The learned parameters also exhibit measurable relationships with speed, acceleration, path length, and displacement, and the magnitude of the activation-weighted angular frequency increases from slow to fast motion. Because $A$, $k$, $\omega$, and $\theta$ are jointly optimized within one response, the present WFB implementation provides a structured feed-forward representation of state and interval. The decoder ablation further indicates that this representation can support a lightweight linear readout, although the preliminary efficiency result requires multi-seed and runtime confirmation.

The backpropagation comparison further clarifies how WFB can be used effectively. A Laplacian-only temporal correction is not aligned sufficiently with the supervised trajectory objective, whereas STD-Laplacian-WFB-FFN preserves task-directed learning and introduces wave curvature as a weak complementary signal. Its best setting, $\lambda=10^{-5}$, improves ADE over STD-WFB-FFN, indicating that local curvature can refine the learned wave field when it is balanced with the standard gradient. Together, the accuracy, capacity-controlled, sensitivity, interpretability, and decoder results establish the intended proof of concept: a learnable wave response can serve as an effective structured component in an otherwise feed-forward predictor.

\section{Conclusion and Future Work}\label{con}

This paper introduced Wave Function Backpropagation (WFB), a wave-parameterized learning formulation that directly associates a neural input with an elapsed temporal interval through learnable amplitude, wavenumber, angular frequency, and phase. Standard task gradients and a spatial-Laplacian curvature regularizer were derived for the wave parameters. The formulation was evaluated in a deliberately controlled feed-forward setting using trajectory prediction as a validation task rather than as the scope of the proposed method.

STD-WFB outperformed the original FFN baseline in both the motion-feature and position-only evaluations. The new capacity-controlled experiment further showed that real-interval WFB is competitive with parameter-matched ReLU predictors, while shuffled-interval WFB achieves the lowest mean ADE. The constant-interval and matched-baseline results indicate that model capacity and the wave representation both contribute to performance, whereas correct interval alignment is not established as the source of the gain. These bounded findings support the intended conclusion: WFB is a feasible and effective structured feed-forward learning formulation.

Future research will investigate optimization stability, parameter identifiability, theoretical approximation properties, computational complexity, tuned periodic baselines, and additional continuously evolving or irregularly sampled data. Sequence-aware propagation is reserved for future work and is not part of the present proof-of-concept evaluation.

\bibliographystyle{unsrt}  
\bibliography{software}

@article{wu2018spatio,
  title={Spatio-temporal backpropagation for training high-performance spiking neural networks},
  author={Wu, Yujie and Deng, Lei and Li, Guoqi and Zhu, Jun and Shi, Luping},
  journal={Frontiers in Neuroscience},
  volume={12},
  pages={331},
  year={2018},
  publisher={Frontiers},
  doi={10.3389/fnins.2018.00331}
}

@article{Ellenberger2025,
  author = {Ellenberger, Benjamin and Haider, Paul and Benitez, Federico and Jordan, Jakob and Max, Kevin and Jaras, Ismael and Kriener, Laura and Petrovici, Mihai A.},
  title = {Backpropagation through space, time and the brain},
  journal = {Nature Communications},
  year = {2025},
  month = {Dec},
  volume = {17},
  number = {1},
  pages = {66},
  doi = {10.1038/s41467-025-66666-z},
  url = {https://doi.org/10.1038/s41467-025-66666-z},
  issn = {2041-1723}
}

@misc{ma2025spikingneuralnetworkstemporal,
      title={Spiking Neural Networks for Temporal Processing: Status Quo and Future Prospects}, 
      author={Chenxiang Ma and Xinyi Chen and Yanchen Li and Qu Yang and Yujie Wu and Guoqi Li and Gang Pan and Huajin Tang and Kay Chen Tan and Jibin Wu},
      year={2025},
      eprint={2502.09449},
      archivePrefix={arXiv},
      primaryClass={cs.NE},
      url={https://arxiv.org/abs/2502.09449}, 
}

@article{Rumelhart1986,
  author  = {Rumelhart, David E. and Hinton, Geoffrey E. and Williams, Ronald J.},
  title   = {Learning representations by back-propagating errors},
  journal = {Nature},
  year    = {1986},
  volume  = {323},
  number  = {6088},
  pages   = {533--536},
  doi     = {10.1038/323533a0}
}

@article{raissi2019physics,
  title={Physics-informed neural networks: A deep learning framework for solving forward and inverse problems involving nonlinear partial differential equations},
  author={Raissi, Maziar and Perdikaris, Paris and Karniadakis, George E},
  journal={Journal of Computational Physics},
  volume={378},
  pages={686--707},
  year={2019},
  publisher={Elsevier}
}

@inproceedings{lerner2007crowd,
  title={Crowds by example},
  author={Lerner, Alon and Chrysanthou, Yiorgos and Lischinski, Dani},
  booktitle={Computer Graphics Forum},
  volume={26},
  number={3},
  pages={655--664},
  year={2007},
  organization={Wiley Online Library}
}

@article{kotseruba2016jaad,
  title={Joint attention in autonomous driving (JAAD)},
  author={Kotseruba, Iuliia and Rasouli, Amir and Tsotsos, John K},
  journal={arXiv preprint arXiv:1609.04741},
  year={2016},
  url={https://arxiv.org/abs/1609.04741}
}

@inproceedings{song2024motion,
  title={Motion Forecasting in Continuous Driving},
  author={Song, N. and others},
  booktitle={NeurIPS},
  year={2024}
}

@article{enhanced2024state,
  title={Enhanced Prediction of Multi-Agent Trajectories via Control Variable Modeling},
  author={Anonymous},
  journal={arXiv},
  year={2024}
}

@inproceedings{bae2024singulartrajectory,
  title={SingularTrajectory: Universal Trajectory Prediction},
  author={Bae, Inhwan and others},
  booktitle={CVPR},
  year={2024}
}

@article{wang2024mdstf,
  title={MDSTF: A Multi-Dimensional Spatio-Temporal Feature Fusion Trajectory Prediction Model for Autonomous Driving},
  author={Wang, Xing and Wu, Zixuan and Jin, Biao and Lin, Mingwei and Zou, Fumin and Liao, Lyuchao},
  journal={Complex \& Intelligent Systems},
  volume={10},
  pages={6647--6665},
  year={2024}
}

@phdthesis{zhang2024pedestrian,
  title={Pedestrian Behavior Prediction Using Machine Learning Methods},
  author={Zhang, Chi},
  school={University of Gothenburg and Chalmers University of Technology},
  year={2024}
}

@inproceedings{gupta2018socialgan,
  title={Social GAN: Socially Acceptable Trajectories with Generative Adversarial Networks},
  author={Gupta, Agrim and Johnson, Justin and Fei-Fei, Li and Savarese, Silvio and Alahi, Alexandre},
  booktitle={Proceedings of the IEEE Conference on Computer Vision and Pattern Recognition},
  year={2018}
}

@article{che2018grud,
  title={Recurrent Neural Networks for Multivariate Time Series with Missing Values},
  author={Che, Zhengping and Purushotham, Sanjay and Cho, Kyunghyun and Sontag, David and Liu, Yan},
  journal={Scientific Reports},
  year={2018}
}

@inproceedings{rubanova2019latentode,
  title={Latent ODEs for Irregularly-Sampled Time Series},
  author={Rubanova, Yulia and Chen, Ricky T. Q. and Duvenaud, David},
  booktitle={Advances in Neural Information Processing Systems},
  year={2019}
}

@inproceedings{kidger2020neuralcde,
  title={Neural Controlled Differential Equations for Irregular Time Series},
  author={Kidger, Patrick and Morrill, James and Foster, James and Lyons, Terry},
  booktitle={Advances in Neural Information Processing Systems},
  year={2020}
}

@INPROCEEDINGS{7780479,
  author={Alahi, Alexandre and Goel, Kratarth and Ramanathan, Vignesh and Robicquet, Alexandre and Fei-Fei, Li and Savarese, Silvio},
  booktitle={2016 IEEE Conference on Computer Vision and Pattern Recognition (CVPR)}, 
  title={Social LSTM: Human Trajectory Prediction in Crowded Spaces}, 
  year={2016},
  volume={},
  number={},
  pages={961-971},
  doi={10.1109/CVPR.2016.110}}

@inproceedings{sitzmann2020implicit,
  title={Implicit Neural Representations with Periodic Activation Functions},
  author={Sitzmann, Vincent and Martel, Julien N. P. and Bergman, Alexander W. and Lindell, David B. and Wetzstein, Gordon},
  booktitle={Advances in Neural Information Processing Systems},
  year={2020}
}

@inproceedings{tancik2020fourier,
  title={Fourier Features Let Networks Learn High Frequency Functions in Low Dimensional Domains},
  author={Tancik, Matthew and Srinivasan, Pratul P. and Mildenhall, Ben and Fridovich-Keil, Sara and Raghavan, Nithin and Singhal, Utkarsh and Ramamoorthi, Ravi and Barron, Jonathan T. and Ng, Ren},
  booktitle={Advances in Neural Information Processing Systems},
  year={2020}
}

\end{document}